\documentclass[preprint,12pt,authoryear, nonatbib]{elsarticle}

\usepackage{url}

\usepackage{times}
\usepackage{epsfig}
\usepackage{graphicx}
\usepackage{amssymb}
\usepackage{booktabs}
\usepackage{multicol}
\usepackage{multirow}
\usepackage{stfloats}
\usepackage{balance}
\usepackage{subfig}
\usepackage{amsmath}
\usepackage[margin=0.75in]{geometry}
\usepackage{apacite}

\begin{document}

\begin{frontmatter}



\title{Impact of Data Breadth and Depth on Performance of Siamese Neural Network Model: Experiments with Three Keystroke Dynamic Datasets}


\author[1]{Ahmed Anu Wahab}
\author[1,4]{Daqing Hou}
\author[2]{Nadia Cheng}
\author[3]{Parker Huntley}
\author[5]{Charles Devlen}

\affiliation[1]{organization={Electrical and Computer Engineering, Clarkson University},
            city={Potsdam},
            state={NY},
            country={USA}}

\affiliation[2]{organization={Statistics, University of Virginia},
            city={Charlottesville},
            state={VA},
            country={USA}}

\affiliation[3]{organization={College of Computing, Georgia Tech},
            city={Atlanta},
            state={GA},
            country={USA}}

\affiliation[4]{organization={Software Engineering, RIT},
            city={Rochester},
            state={NY},
            country={USA}}

\affiliation[5]{organization={Computing and Information Science, RIT},
            city={Rochester},
            state={NY},
            country={USA}}

\begin{abstract}
Deep learning models, such as the Siamese Neural Networks (SNN), have shown great potential in capturing the intricate patterns in behavioral data. However, the impacts of dataset breadth (i.e., the number of subjects) and depth (e.g., the amount of training samples per subject) on the performance of these models is often informally assumed, and remains under-explored. To this end, we have conducted extensive experiments using the concepts of ``feature space'' and ``density'' to guide and gain deeper understanding on the impact of dataset breadth and depth on three publicly available keystroke datasets (Aalto, CMU and Clarkson II). Through varying the number of training subjects, number of samples per subject, amount of data in each sample, and number of triplets used in training, we found that when feasible, increasing dataset breadth enables the training of a well-trained model that effectively captures more inter-subject variability. In contrast, we find that the extent of depth's impact from a dataset depends on the nature of the dataset. Free-text datasets are influenced by all three depth-wise factors; inadequate samples per subject, sequence length, training triplets and gallery sample size, which may all lead to an under-trained model. Fixed-text datasets are less affected by these factors, and as such make it easier to create a well-trained model. These findings shed light on the importance of dataset breadth and depth in training deep learning models for behavioral biometrics and provide valuable insights for designing more effective authentication systems.
\end{abstract}

\end{frontmatter}


\section{Introduction}
Keystroke dynamics is an emerging authentication solution that leverages a user's behavioral patterns extracted from typing data for identity verification and/or identification. These patterns have been found to be quite unique to each user and can be used for authentication. An instance of keystroke data is comprised of timestamps and key names for every key-press and key-release event. From this raw data, features like monographs and digraphs are derived using the timing information of the keys. Monographs refer to individual key events, while digraphs represent pairs of consecutive key events.
Main advantages of keystroke dynamics include passiveness, unobtrusiveness, and cost-effectiveness (requiring no additional hardware), making it an attractive solution. Moreover, unlike other methods, behavioral biometrics, such as keystroke dynamics, can continuously monitor a user's behavioral pattern for anomalies and prevent account takeovers beyond the login point, which is known as continuous authentication.

In keystroke dynamics, traditional methods for verifying users' identities involves the use of statistical algorithms such as distance/similarity measures, cluster analysis, and probability measures by analysis of keystroke timings \cite{teh2013survey, wahab2022securing, ayotte2020fast, gunetti2005keystroke, killourhy2009comparing}. While the adoption of deep learning models built upon Artificial Neural Networks has been expanding in recent years \cite{harun2010performance, andrean2020keystroke, deng2013keystroke}, these models are often binary classifiers, which require a model for each subject. This makes binary classifier solutions difficult to scale, as they demand a substantial volume of data per subject to train adequately, as well as substantial amounts of storage.

To overcome these limitations, the Siamese Neural Network (SNN) was introduced for keystroke dynamics \cite{typenet}. SNN was originally implemented for image classification and person re-identification \cite{bromley1993signature}, and is specifically designed for comparing the similarity between two or more inputs. A SNN trains two or more identical sub-networks, each of which produces an output vector known as an embedding, which is a lower-dimensional representation of the input vector. These embeddings are then compared to produce a similarity score between the inputs, allowing the model to compute similarity scores even for new subjects that were never seen during training, making it an effective and scalable model in keystroke dynamics. Furthermore, SNNs address the imbalanced class data problem, as they distinguish between inputs as either similar or dissimilar during training. While it is generally known that deep learning models like SNNs require large amounts of data to train, there are no guidelines available on exactly how much data is needed to train them, or how their performance is affected by factors like the number of subjects or the number of samples per subject within the dataset.


\begin{table*}[tb]
\begin{center}
\scriptsize
\renewcommand*{\arraystretch}{1.2}
\begin{tabular}{l l l}
\toprule
\textbf{Keystroke Dataset} & \textbf{\#Subjects}  & \textbf{Data per subject}\\
\midrule
 CMU \cite{killourhy2009comparing} & 51 &  `.tie5Roanl' 400 times\\
 GreyC (A) \cite{giot2009greyc} & 133 & `greyc laboratory' 51 times\\
 GreyC (B) \cite{giot2012web} & 83 & ~132 samples\\
 Clarkson I \cite{vural2014shared} & 39 & 21,533 characters\\
 Clarkson II \cite{murphy2017shared} & 103 & 125,000 keystrokes\\
 Buffalo \cite{sun2016shared} & 148 & 17,000 keystrokes\\
 Account Recovery \cite{wahab2021utilizing} & 44 & 5,609 characters\\
 Multi-Keyboard \cite{wahab2022shared} & 60 & 14,000 keystrokes\\
 CU Multi-modality  \cite{ray2023multi} & 88 & 4,391 characters\\
 Aalto Mobile \cite{palin2019people} & 37,370 & 15 sentences\\
 Aalto Desktop \cite{dhakal2018observations} & 168,000 & 15 sentences\\
\bottomrule
\end{tabular}
\end{center}
\caption{Publicly available keystroke behavioral biometric datasets,  numbers of subject, and amount of data per subject.}
\label{public_datasets}
\end{table*}

\begin{figure}[tb]
\begin{center}
\includegraphics[width=0.75\linewidth]{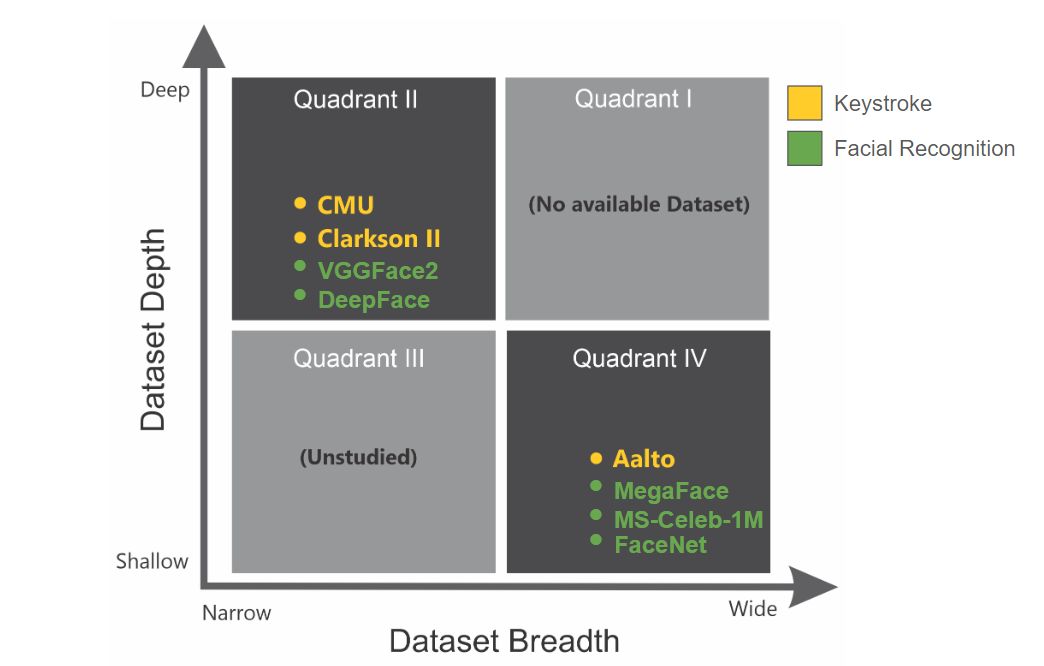}
\end{center}
   \caption{Quadrant plots of datasets based on breadth and depth}
\label{breadth_vs_depth_quadrant}
\end{figure}
 
As shown in Figure \ref{breadth_vs_depth_quadrant}, the breadth of a dataset, defined by the number of subjects it contains, can be either narrow or wide, and depth, defined by the amount of data per subject, can be shallow or deep. This results in four possible combinations: wide and deep, narrow and deep, narrow and shallow, and wide and shallow. To the best of our knowledge, there are currently no keystroke dynamics datasets falling in Quadrant I, likely because datasets that are both broad and deep are expensive to collect. The CMU and Clarkson II datasets fall under Quadrant II, representing narrow but deep datasets, while the Aalto dataset occupies Quadrant IV, representing a wide but shallow dataset. Quadrant III, signifying narrow and shallow datasets, is not included in our study due to the exceedingly small dataset sizes that would be present.

Facial recognition is a long studied biometric technology that in recent years has shifted from traditional statistical and ML methods to deep learning methods \cite{wang2021deep}. Wang et al. \cite{wang2021deep} surveyed deep face recognition datasets, and classified them using the same terms as this paper, breadth and depth. The publicly available datasets they highlighted as being high in breadth and depth are VGGFace2 \cite{cao2018vggface2}, MS-Celeb-1M \cite{guo2016ms}, and Megaface \cite{megaface}. The VGGFace2 dataset contains 9,131 subjects, and an average of 362.6 photos per subject, making it belong to the depth category. Conversely, Wang et al. classified the MS-Celeb-1M and MegaFace datasets as belonging to the breadth category, with 100,000 subjects for MS-Celeb-1M and 672,057 subjects for MegaFace, with a mean depth of 100 and 7 images per subject respectively. According to our taxonomy, this would put VGGFace2 in Quadrant II, and MS-Celeb-1M and MegaFace in Quadrant IV, as reflected in Figure \ref{breadth_vs_depth_quadrant}. While Aalto and MegaFace are both in Quadrant IV, both being wide but shallow, MegaFace is still much higher in breadth than the Aalto dataset, at 672,057 subjects compared to 168,000 subjects. As Aalto is the current broadest keystroke dynamics dataset, this indicates that to advance research in this area, larger scale datasets are needed. However, more work is needed to determine whether these datasets should be higher in breadth, depth, or both.

As facial recognition moved to deep methods, the need for large scale data for training these models has grown. Zhang et al. \cite{zhang2017} examined the long tail effect in deep FR model training data, which is inherent in facial datasets sourced from the web that contain classes with widely varying amounts of depth. Using the MS-Celeb-1M dataset, Zhang et al. curated a tailed training set, where any subject with less than 20 images was considered tailed data. They found that training on exclusively tailless users doesn't always give the best performance, and the inclusion of 20-50\% of tailed data increased accuracy 0.1 to 0.12\%. From these findings, it is clear that training set depth has a significant impact on the accuracy of FR models, but the question of whether depth or breadth can compensate for the lack of the other is not sufficiently explored.

FaceNet \cite{facenet} and DeepFace \cite{deepface} are two formerly state-of-the-art facial recognition models that both utilized extremely large-scale training datasets. FaceNet was trained on approximately 8 million identities, with 50 images per subject, while DeepFace was trained with approximately 4 thousand identities with between 800-1200 images per subject. According to our taxonomy shown in Figure \ref{breadth_vs_depth_quadrant}, FaceNet belongs to Quadrant IV, while the DeepFace belongs to Quadrant II. While FaceNet and DeepFace both achieve high accuracy in facial recognition tasks despite belonging to different quadrants, no concrete conclusions on the effect of breadth and depth on model performance can be drawn due to their differing architectures and alignment methods. Moreover, experimentation with these large-scale datasets is not possible due to their private nature.

In order to more conclusively examine the effects of breadth and depth on generalizability and performance, we utilize the same state-of-the-art TypeNet architecture trained on 3 publicly available datasets with varying degrees of breadth and depth. The unified architecture used in our experiments allows us to pin down the effect of breadth and depth on model performance, and the use of 3 publicly available datasets allows the findings to be replicable by other researchers in this field. While facial recognition can leverage ``in the wild” data from the web, as in the case of Labeled Faces in the Wild \cite{LFWTech, LFWTechUpdate}, keystroke dynamics data can not be obtained in this manner, and typically takes longer to collect, making large-scale keystroke dataset collection expensive. The findings of our paper will be of crucial importance for helping researchers in this field collect keystroke datasets in a better informed manner, so time and expense are appropriately spent to help develop large-scale datasets that improve the current state-of-the-art. The lack of a clear understanding of the amount and nature of data needed for training effective Siamese Neural Networks hinders the development of keystroke dynamics. 

Guided by this taxonomy of datasets, we have conducted experiments using the Aalto \cite{dhakal2018observations}, CMU \cite{killourhy2009comparing}, and Clarkson II \cite{murphy2017shared} keystroke datasets, which also include both fixed-text and free-text (both controlled and uncontrolled) categories of keystroke dynamics. Our goal was to investigate the impact of breadth and  depth of a dataset on the performance of an SNN model. To guide our investigations, we proposed a theory around the notions of ``\textit{feature space}'' and ``\textit{density}'' to further understand this impact, and conducted experiments to validate the theory.

The \textit{feature space} represents the range of unique features, including graphs, timing information, and other identifiers, that the network learns to distinguish between subjects. It encompasses the set of characteristics and patterns used by the network for subject differentiation. \textit{Density}, in this context, refers to how concentrated or scattered the data points are for each feature  within the feature space. ``Large density''  signifies a scenario where there are many data points, or similar graphs, closely packed within each unit in the feature space, potentially leading to better model performance.

We trained with different subsets of the Aalto dataset (breadth-wise and depth-wise) to determine the optimal number of subjects, training size, amount of data per subject, and sample sequence length, for achieving high performance with SNN. Furthermore, the CMU and Clarkson II were employed to further delve deeper into the impact of dataset depth, given their larger number of samples per subject but fewer subjects in total (i.e., Quadrant II in Figure \ref{breadth_vs_depth_quadrant}). 

In the model evaluation phase, a consistent performance improvement as the gallery sample size ($G$) increases indicates a sufficiently trained model. Another characteristics of a sufficiently trained model is its ability to maintain consistent and reliable performance even after re-sampling or with the introduction of additional data in terms of its breadth (number of subjects) or depth (amount of data per subject). This stability in performance results indicates that the model has reached a point where further data inclusion does not produce substantial improvements. In contrast, a model with fluctuating performance as $G$ increases signifies an under- or inadequately-trained model. Such an under-trained model is often the result of having an extensive feature space but an insufficient number of training triplets to learn the whole feature space. Similarly, a model that shows significant performance variations after re-sampling or when additional data is introduced is also considered under-trained.

Our experiments provided valuable insights into the role of dataset breadth and depth in determining the performance of Siamese networks-based behavioral biometric. Our findings showed that both dataset breadth and depth impact performance, with these effects predominantly observed in free-text datasets.
Furthermore, leveraging the concept of ``feature space'' and ``density'', the results provide insights into the critical aspect of determining the levels of performance that can be expected of Siamese networks based on the dataset's breadth and depth. They also provide guidance on the level of performance improvement that can be achieved when more data is added to the dataset.

This work has made the following contributions:
\begin{itemize}
    \item We investigate the impact of dataset breadth and depth on the performance of an SNN deep learning model for keystroke dynamics authentication.
    Our extensive experimentation uses both fixed-text and free-text datasets, including both controlled and uncontrolled environments, which constitute the two primary categories of keystroke dynamic.
    

    \item We showed that by increasing the breadth of the training dataset (the case of Aalto), a well-trained model can be created, for which the impact of depth is reduced and even diminishes. On the other hand, when the breadth of the training dataset is too narrow, the model performance is substantially impacted by the depth. 

    \item Our experiments demonstrate that both CMU and Aalto datasets can be used to produce a fully and sufficiently trained model. On the other hand, the Clarkson II dataset, with its extensive feature space due to its free-text nature, requires a significantly larger number of training triplets. The models for Clarkson II remain under-trained even after being trained with as many as 15.2 and 30.4 million triplets.
    Generating additional training triplets is necessary to achieve a well-trained model for this dataset.

    \item We established a new state-of-the-art performance of 0.7\% EER for the CMU dataset, a 76\% performance improvement over previous work \cite{deng2013keystroke, maheshwary2017deep},
    despite having 5 times less gallery sample data.

    \item We provide valuable insights into the potential performance improvements attainable through an increase in dataset size, either in terms of breadth or depth.
    
\end{itemize}

A preceding version of this article was presented in \cite{wahab2023impact}. This paper makes significant advancements over the preliminary work in the following areas:
\begin{itemize}
    \item We experimented with two additional keystroke datasets (CMU \cite{killourhy2009comparing} and Clarkson II \cite{murphy2017shared}), representing the fixed-text and uncontrolled free-text categories, respectively.
    \item We add experiments to further delve into the impact of dataset depth in relation to: (i) the number of samples per subject, (ii) the sequence length, represented by the number of data in each sample, and (iii) the total number of training triplets for training.
    \item We used the notions of ``feature space'' and ``density'' to further explain how dataset depth affects the performance of Siamese networks.
    \item We compared our results from the two additional datasets with the state-of-the-art results and methods.
\end{itemize}

The rest of the paper is organized as follows. Section 2 surveys related work. Sections 3, 4, 5 and 6 describe the Siamese neural network architecture, the datasets, the experimental procedures, and the results, respectively. Section 7 concludes the paper.

\section{Related Work}
Behavioral biometrics, particularly in keystroke dynamics, have used simple distance classifier and outlier detection methods. These approaches typically involve capturing various features such as monographs, which are single keystrokes from individual characters; and digraphs, which capture the relationship between two consecutive keystrokes, such as \textit{down-down}, \textit{down-up}, \textit{up-down}, and \textit{up-up}. Commonly used distance classifiers and outlier detection methods include the Manhattan distance, euclidean distance, Mahanalobis distance, k-nearest neighbours, k-means clustering, and their  variants \cite{killourhy2009comparing, ayotte2020fast, wahab2022securing, zhong2012keystroke, zhong2015survey}. 


The distance and outlier detection techniques have demonstrated reasonable performance in keystroke dynamics authentication. However, they rely heavily on the extracted features and require strong domain knowledge for manual feature engineering. With the advent of deep learning, the trend has shifted towards utilizing deep neural networks for keystroke biometrics. Deep learning offers several advantages over the traditional approaches and have the ability to automatically extract relevant features from raw keystroke data without the need for explicit or extensive feature engineering. Harun et al. \cite{harun2010performance}, did a comprehensive comparison between two artificial neural networks (ANN) and several distance-based classifiers on four different datasets. All four datasets have a total of 47 subjects combined, which are small. 
They showed that the ANNs (MLP and RBF) are more suitable to discriminate and classify nonlinear keystroke data and reported EER as low as 2\%. Andrean et al. \cite{andrean2020keystroke} also used an MLP network on a fixed-text keystroke dataset known as the CMU dataset \cite{killourhy2009comparing} and reported a 4.45\% EER which outperformed the results from the original benchmark classifiers.
Deng and Zhong \cite{deng2013keystroke} created a Deep Belief Net (DBN), which are probabilistic generative models that are composed of multiple layers of hidden variables. The DBN, a binary classifier, was individually trained for each subject in the CMU dataset using their initial 200 samples from the genuine subject's data. Testing was conducted using the remaining 200 samples, along with samples from other subjects serving as impostor samples. This approach yielded an EER of 3.5\%, further showing an improvement over the benchmark results. Maheshwary et al. \cite{maheshwary2017deep} achieved a slightly improved performance with an EER of 3\% on the CMU dataset, employing a similar approach as in \cite{deng2013keystroke}, but utilizing a feed-forward neural network model referred to as Deep Secure.

\begin{figure*}[!t]
\centering
\subfloat[\centering Sub-network Architecture]{
\includegraphics[width=0.48\textwidth]{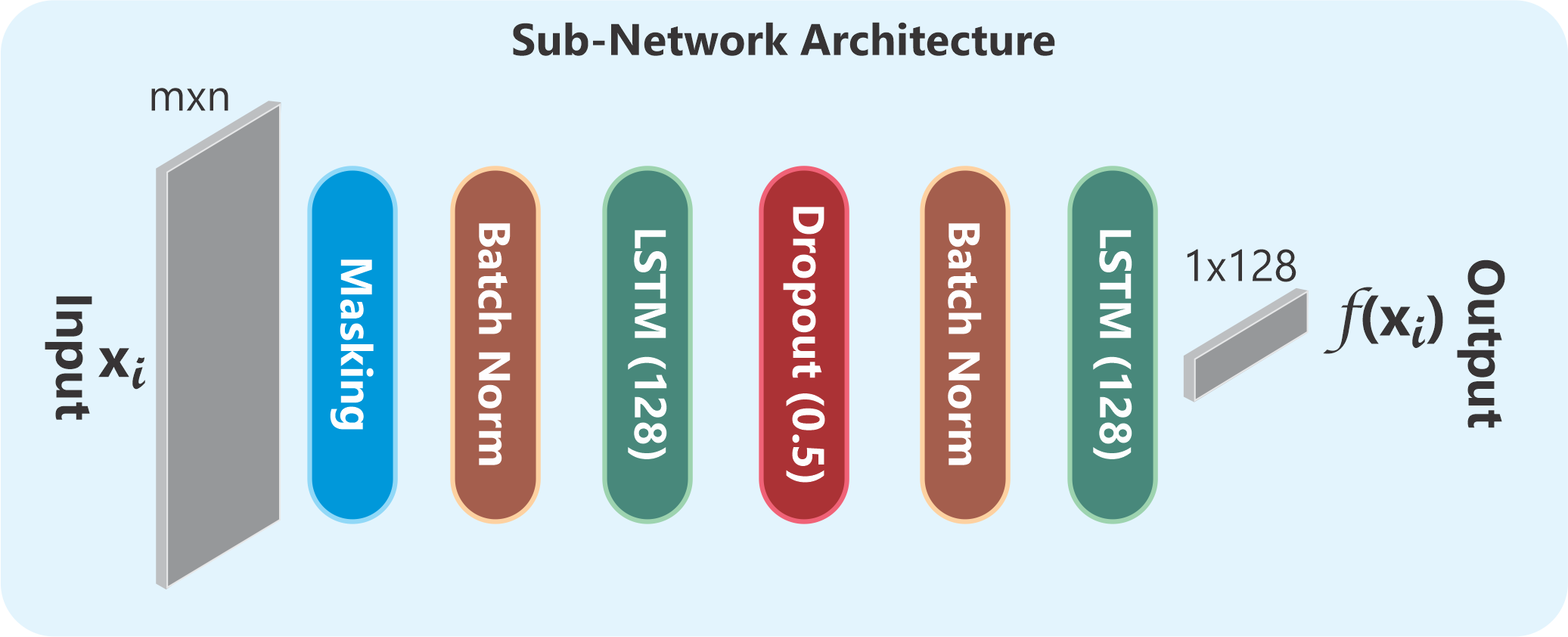}%
\label{fig_sub-network_arch}}
\hfil
\subfloat[Siamese Network Architecture]{\includegraphics[width=0.48\textwidth]{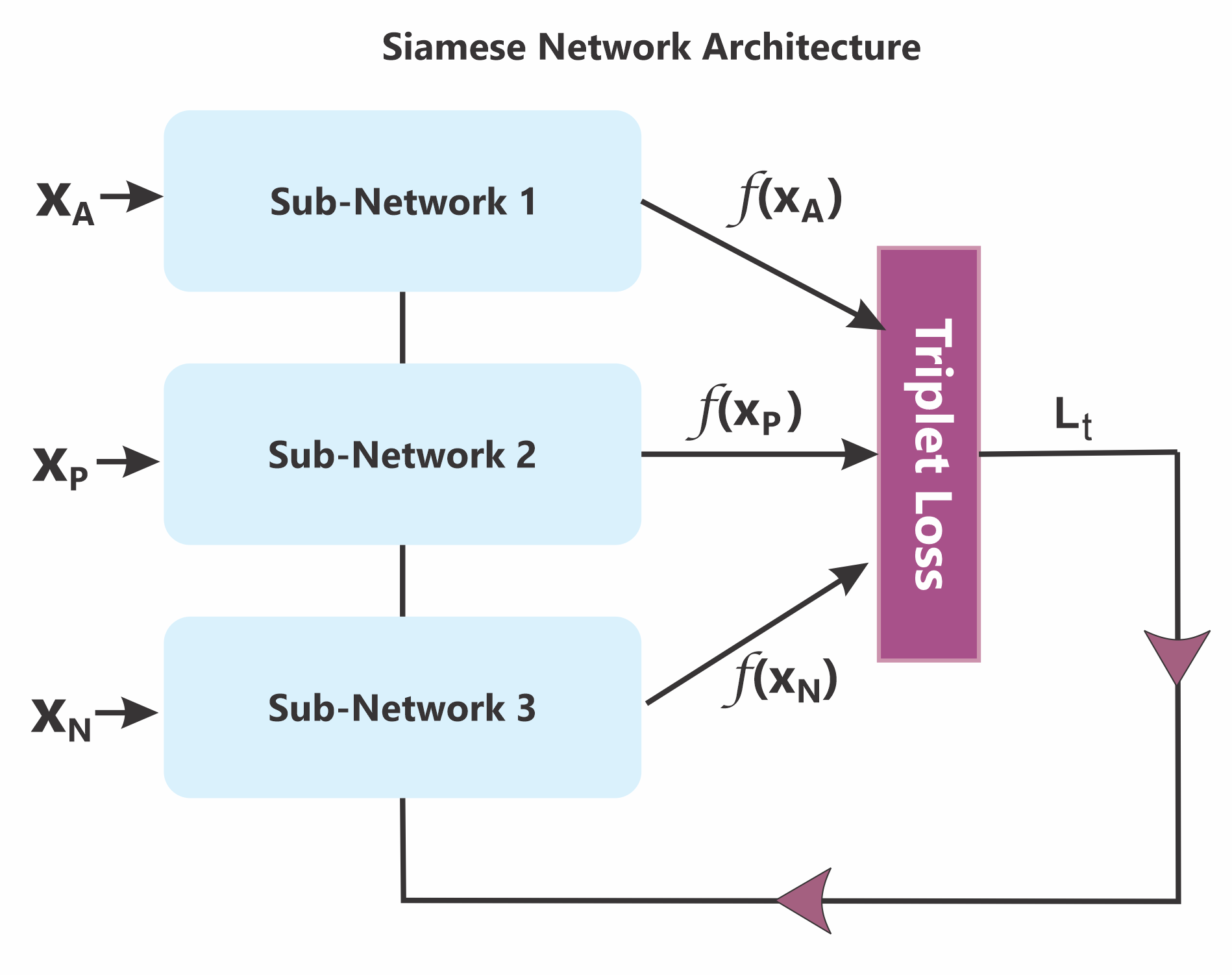}%
\label{fig_network_arch}}
\caption{(a) The Siamese sub-network, taking a time series input (\textbf{x}\textsubscript{$i$}) of shape $m\times n$ and returning an output vector (embeddings) of shape $1\times 128$. (b) The Siamese network, consisting of three (3) sub-networks. Loss  is calculated from the three output vectors and are back-propagated into the  network.}
\label{fig_snn_arch}
\end{figure*}

Although these early work on the application of deep learning for keystroke dynamics achieved better performance compared to the traditional distance-based or outlier detection methods, they were trained and tested on small datasets. Since the evolution of deep learning, it is common knowledge that datasets with large data sizes are required to effectively train deep learning models and capture complex temporal dependencies present in keystroke data. However,  as shown in Table \ref{public_datasets}, most publicly available keystroke datasets are very limited in size, typically with a few tens or hundreds of subjects. These small datasets have limited the full exploration of deep learning in keystroke dynamics until 2018 when the Aalto dataset \cite{dhakal2018observations} was released. The Aalto dataset has collected 136 million keystrokes data  from 168,000 subjects.

Therefore, our work is needed to quantify the breadth and depth of keystroke data needed in order to quantify how breadth and depth affects generalizability and performance of state-of-the-art keystroke dynamics deep learning models. 


Acien et al. \cite{typenet} developed TypeNet, an SNN architecture based on Long Short-Term Memory (LSTM) networks. Their model, trained with 68,000 subjects and tested with 1,000 subjects from the Aalto dataset \cite{dhakal2018observations}, achieved an EER of 1.2\%.
Their work showed a significant improvement over past work as it leveraged a large amount of data and was tested on a large number of subjects unseen during training, making it more realistic. 
Nevertheless, their study did not explore the performance of SNN on smaller datasets, nor did it investigate the impact of datasets' breadth and depth on wide and shallow, or narrow and deep datasets. It's worth noting that Acien et al. conducted cross-dataset testing by training a model on the Aalto dataset and evaluating its performance on other free-text datasets, such as the Clarkson II dataset \cite{vural2014shared}. However, the results from these cross-dataset tests (26.8\% EER) fell considerably short of the state-of-the-art performance of 7.9\% EER achieved by Ayotte et al. \cite{ayotte2019fast} on the Clarkson II dataset with 50 sequence length. 
Another study by Killhoury and Maxion \cite{killourhy2010did} examined six factors that could affect anomaly detector performance on the CMU dataset. The effect of algorithm, amount of training data, feature set, model data updating, impostor practice, and typist-to-typist variation on performance were evaluated on three different classifiers. They found that algorithm, amount of training data, and model data updating had the largest effect on detector performance. However, their experiments were only performed on the CMU dataset and with a small 15 user validation set, and did not quantify the effect of breadth.
Our work is therefore novel in that it addresses these gaps in the literature, as no previous studies have explored the impact of dataset breadth and depth on deep learning performance in keystroke dynamics or any other behavioral biometric modality.

\section{The Siamese Neural Network Architecture}
SNN is used to find the similarity between inputs by comparing the output vectors (embeddings) of the sub-networks. Figure \ref{fig_sub-network_arch} shows the architecture of the Siamese sub-network, which includes several layers designed to optimize the performance of the model. First, there is a masking layer that helps prevent the model from training on zero-padded rows, which are added if the input has fewer rows than the desired sequence length ($m$). Hence, the zero-padded rows do not contribute to the computed loss value. Next, batch normalization layers are applied to normalize the input data and improve the training speed and stability. Two LSTM layers follow, which are activated using the hyperbolic tangent function (tanh) to capture the temporal dependencies in the sequential data. Lastly, a dropout layer is used as a regularization technique to prevent overfitting.

The SNN architecture in Figure \ref{fig_network_arch} consists of three (triplet) sub-networks that share weights and are trained together to learn meaningful representations of input data. Each sub-network takes in a single input (\textbf{x}\textsubscript{$i$}) and produces an output vector (\textbf{f}\textbf{x}\textsubscript{$i$}). The first sub-network takes in an anchor sample (\textbf{x}\textsubscript{$A$}) and produces a vector representation of it (\textbf{f}\textbf{x}\textsubscript{$A$}). The second sub-network takes in a positive sample (\textbf{x}\textsubscript{$P$}) and produces another vector representation of it (\textbf{f}\textbf{x}\textsubscript{$P$}). The third sub-network takes in a negative sample (\textbf{x}\textsubscript{$N$}) and produces its own vector representation (\textbf{f}\textbf{x}\textsubscript{$N$}). The anchor and positive samples are drawn from the same subject (i.e., legitimate user's data), while the negative samples are selected from the impostors' data. The three output vectors are then passed through the triplet loss function to update the weights of the entire Siamese network. The triplet loss function trains the network by minimizing the distance between anchor and positive samples while maximizing the distance between anchor and negative samples as shown in Equation \ref{triplet_loss}, where $\alpha$ is a hyperparameter that controls the degree of separation between the anchor and negative samples in the embedding space.

\begin{equation}
\label{triplet_loss}
L_t = \max\{0, ||\mathbf{f}(\mathbf{x}^i_A) - \mathbf{f}(\mathbf{x}^i_P)||^2 - ||\mathbf{f}(\mathbf{x}^i_A) - \mathbf{f}(\mathbf{x}^j_N)||^2 + \alpha\}
\end{equation}


\section{Datasets}
\label{sec:datasets}

The keystroke datasets used for our experiments are  Aalto \cite{dhakal2018observations}, CMU \cite{killourhy2009comparing} and Clarkson II \cite{murphy2017shared}, which are publicly available.

\subsection{The Aalto Dataset}
The Aalto University desktop dataset~\cite{dhakal2018observations} is a large-scale controlled free-text dataset collected using an online typing test on desktop computers. The dataset has 136 million keystrokes collected from 168,000 subjects and for a duration of three months, each subject transcribing 15 English sentences which were randomly drawn from a set of 1,525 examples consisting of at least 3 words, and a maximum of 70 characters per sentence.
The characters typed can exceed 70 as subjects are allowed to make typing errors, correct them or add new characters when typing. As a result of the large number of subjects and limited samples per subject, this dataset is wide but shallow.


\subsection{The CMU Dataset}
The CMU dataset \cite{killourhy2009comparing}, one of the extensively analyzed fixed-text datasets, comprises keystroke data from 51 subjects who repeatedly typed a predefined static password string ``.tie5Roanl'', 50 times in each of 8 sessions, with atleast one day between sessions, resulting in a total of 400 password-typing samples per subject. Participants were required to input the ten-character password accurately in a sequential order, followed by the ``Enter'' key. In cases of sequencing errors, participants reentered the password.

\subsection{The Clarkson II Dataset}
The Clarkson II dataset \cite{murphy2017shared}, collected over a 2.5-year period from 103 subjects, is a distinctive, fully uncontrolled free-text dataset consisting of both keystrokes and mouse interactions. The logger was installed on participants' computers, unobtrusively capturing non-sensitive keystroke data to a remote server. This dataset is notably different for its uncontrolled nature, capturing the full spectrum of participants' computer activities, including gaming, making it a valuable real-world representation of diverse user interactions. Each subject contributed an average of 125,000 keystrokes, resulting in a combined total of 12.9 million keystrokes.

\subsection{Data Preprocessing}
To enhance the SNN's ability to learn relevant information from the data more efficiently, we performed data preprocessing on the datasets.

\begin{figure}[tb]
\begin{center}
\includegraphics[width=0.5\linewidth]{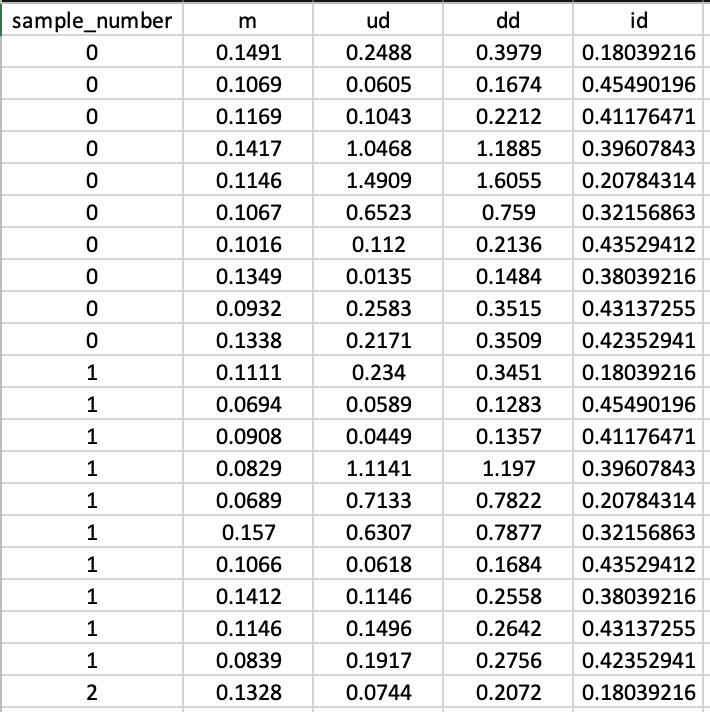}
\end{center}
   \caption{A screenshot of the preprocessed CMU dataset highlighting its four distinctive features (m, ud, dd, and id).}
\label{sample_data}
\end{figure}

With the exception of the CMU dataset, which had already undergone preprocessing and included three time-features (m, ud, and dd), we extracted four time-features (monographs and digraphs) from the remaining keystroke datasets. These features are \textit{m, ud, dd, uu}, where \textit{m} is the duration between the press and release of a single key; \textit{u} and \textit{d} represent \textit{UP (release)} and \textit{DOWN (press)}. Hence, \textit{ud} is the time interval between the release of a key and the press of the next key. The extracted features have duration in milliseconds and ranges between 0 and 1.
In keystrokes, the keys typed are also distinguishable features, so an additional feature, \textit{id} was added to the time-features, where \textit{id} is the ASCII value of key presses/releases divided by 255, which forces the values to range between 0 and 1. Overall, 4 features were extracted from the CMU dataset, while 5 features were extracted from the Aalto and Clarkson II keystroke datasets. Finally, we filtered out any potential outliers in the data by removing rows containing digraphs that exceed 5 seconds. Figure \ref{sample_data} shows the preprocessed CMU dataset, with each row denoting the extracted features between a key and its immediate subsequent key. Each sample consists of ten rows, matching the sequence length ($M$).

\section{Experimental Procedures and Results}
In order to investigate the effect of data size on Siamese networks and determine the most important dimension (breadth or depth) of a dataset for achieving optimal performance, we conducted comprehensive experiments in two categories. The first category is known as ``breadth-wise,'' which examines the performance of SNNs on a large dataset with varying numbers of subjects.
The second category, ``depth-wise,'' examines the performance of SNN based on the dataset depth.

The SNN architecture was implemented using the Tensorflow library on a 24 GB Nvidia GeForce RTX 3090.
Each experiment was conducted using a margin ($\alpha$) of 1.5, which yielded the best performance. The runtime for training one model averaged around 14 hours, and results were reported using the Equal Error Rate (EER) metric.
To ensure the reliability of our results, we repeated each experiment ten times, each time with random subject selection. This approach minimizes any potential variation in the results and obtains a more accurate estimate of the model performance. 
 
\subsection{Breadth-wise Experiments}
We focused our breadth-wise experiments exclusively on the larger dataset: Aalto. This choice stemmed from the datasets' substantial size, enabling us to create distinct subject groups for a more effective investigation.
Using the Aalto dataset, we randomly selected 1,000 out of the 168,000 subjects for testing purposes. 
To ensure that our experiments cover a diverse range of subjects, we randomly selected 10 groups of subjects from the remaining pool, with replacement, for training. The number of subjects in each group are 125, 250, 500, 1,000, 2,000, 4,000, 8,500, 17,000, 34,000, and 68,000. For each group, we created a data generator for generating the required input triplets for the Siamese network. The total number of possible triplets ($T$\textsubscript{possible}) that can be generated from $P$ subjects is calculated as shown in Equation \ref{g_pairs} and \ref{possible_triplets}, where $G$\textsubscript{pairs} is the number of possible sample pairs that can be created from the genuine subject's data. Given that each subject contributed 15 samples, there will be over 24 million and 7.28 trillion possible triplets for 125 and 68,000 subjects respectively. 
Since generating all possible triplets from the dataset would be computationally expensive and memory draining, we randomly generated only 7.6 million triplets, out of the total possible triplets, for training the Siamese network in this category. This number was empirically selected to ensure that the training process was not excessively computationally expensive, while still providing enough triplets to effectively train the model.
Furthermore, with this selection, each group of subjects has ample amount of triplet data required for model training. We maintained a fixed number of samples per subject at 15, a gallery sample size ($G$) of 10, and sequence length ($m$) of 70, representing the maximum number of rows of data in each sample. Samples data exceeding this limit are truncated, while those below it are zero-padded.

\begin{equation}
    G_{\text{pairs}} = \text{No of samples} \times \left(\frac{\text{No of samples} - 1}{2} \right)
    \label{g_pairs}
\end{equation}

\begin{equation}
    T_{\text{possible}} = P \times \Bigl(G_{\text{pairs}} \times (P - 1) \times \text{No of samples} \Bigr)
    \label{possible_triplets}
\end{equation}

\subsection{Depth-wise Experiments}
We examined three factors that provide a more comprehensive understanding of dataset depth's impact on SNN performance. These factors include: (1) varying number of samples per subject, (2) varying sample sequence length, and (3) varying amount of training triplets

To understand the impact of these variations on Siamese network performance, we relied on the concepts of ``\textit{Feature Space}'' and ``\textit{Density}.'' \textit{Feature Space} represents the range of unique features the network learns to distinguish between subjects. This space is influenced by the nature of the dataset, the extent of subject diversity, and the variability of input data. \textit{Density}, on the other hand, relates to how data points are distributed within the feature space. In a dataset with high density, data points or feature representations are concentrated within a limited feature space. In contrast, datasets with low density have data points more scattered across an extensive feature space. These concepts served as our guiding framework for analyzing the influence of dataset depth on Siamese network performance.

\subsubsection{Varying Number of Samples per Subject}
We conducted experiments using different numbers of samples per subject during training. Specifically, with the Aalto (wide and shallow) dataset, where each subject sample represent a sentence, and each subject typed 15 different sentences from a pool of 1525 sentences, using more samples per subject enlarges the feature space, covering a wider range of graphs and timing variations. Similarly, as more samples per subject are added, the density within the feature space also increases due to the accumulation of more similar graphs and timing data points.
Clarkson II, a narrow and deep dataset, consisting of uncontrolled free-text input with no two identical samples, and large intra-variance, inherently possesses a larger and more diverse feature space. Increasing the number of samples per subject further expands this feature space but also results in a more scattered density, as a result of large intra-variance. 
Therefore, we hypothesize that varying the number of samples per subject has an impact on these two datasets.
In contrast, the CMU, a narrow and deep dataset, where subjects typed the same static password repeatedly, has a relatively fixed feature space. Increasing the number of samples per subject does not significantly expand the feature space but instead increases the density since all samples are repetitions of the same text. Therefore, we hypothesize that varying the number of samples per subject would have little to no impact on performance. Multiple experiments were conducted to investigate this.

\subsubsection{Varying Sample Sequence Length}
The sequence length defines the number of rows in each sample and establishes the length of data rows required before an authentication or verification decision is made. Typically, models trained with longer sequences tend to have improved performance, particularly when a substantial number of training triplets are used for training.
Therefore, we hypothesize that varying the sequence length would impact the performance of all three datasets. However, it is important to note that the CMU dataset is a password-based fixed-text dataset with a fixed sequence length of 10. Altering this would contradict the purpose of the dataset, so we did not modify it.

\subsubsection{Varying Amount of Training Triplets}
To reduce computational overhead and account for potential performance plateaus, we often employ a smaller number of training triplets relative to the total possible triplets calculated by Equations 2 and 3 for each dataset. Our hypothesis revolves around the dataset's feature space; larger feature spaces demands more triplets for effective training, as in the case of the Clarkson II dataset with its expansive feature space, given its uncontrolled free-text nature. Conversely, the CMU dataset, with its fixed-text nature and relatively fixed feature space, would demand fewer triplets. To validate this hypothesis, we conducted experiments to examine how varying the number of training triplets impacts performance across all three datasets.

Overall, for the CMU dataset, out of a total of 51 subjects, we reserved 5 subjects for testing, 5 for validation, and used the remaining 41 for training. The Clarkson II dataset, on the other hand, comprises of 103 subjects, but only 72 of them have contributed a minimum of 10,000 keystrokes. So, we assigned 5 of these 72 subjects for testing, 5 for validation, and the remaining for training.

\subsection{Model Evaluation}
The evaluation of all models, including both breadth-wise and depth-wise experiments, was performed using a dedicated test dataset, ensuring no overlap between the training and test subjects. To ensure fairness and unbiased evaluation across all experiments, the same set of test users was consistently utilized for all experiments within each dataset. In the evaluation process, we adhered to a standardized procedure. For each of the $T$ test subjects, we randomly selected samples from each of the remaining $T - 1$ subjects to serve as impostor samples. These impostor samples, along with the genuine samples, make up the subject's test samples. Using the model, we obtained genuine and impostor embeddings for each subject's test samples.
A profile template (also known as gallery) for each subject serves as a reference point for validating new query samples. The first $G$ genuine embeddings  were used as the profile. The remaining ones were utilized as genuine query embeddings $(Q\textsubscript{g})$, with G set to 10 in breadth-wise experiments and varied in depth-wise experiments.
We calculated the pairwise Euclidean distance between the gallery embeddings $(G)$ and the genuine query embeddings $(Q\textsubscript{g})$, resulting in genuine similarity scores. Likewise, we calculated the pairwise Euclidean distance between the gallery embeddings $(G)$ and the impostor embeddings $(Q\textsubscript{i})$, which provided impostor similarity scores. The EER was then computed based on these scores. We report the average EER across all subjects as the final model performance.

\section{Results}
This section presents the results separately for the breadth-wise (Aalto) and depth-wise experiments (Aalto, CMU, and Clarkson II).

\subsection{Results of Breadth-wise Experiments }
For the breadth-wise experiments, only the Aalto dataset can be used based on its large amount of subjects. Table \ref{aalto_breadth_wise_table} shows the average EERs of the SNN models with varying numbers of training subjects from the Aalto dataset. Below are the key observations:
\begin{enumerate}
    \item The EERs reveal a clear pattern of exponential decay, indicating that the performance of the SNN model improves significantly as the number of training subjects increases. For instance, a model trained with 15 samples per subject, and 7.6 million triplets obtained from 125 subjects resulted in an average EER of 7.94\%. However, when the same number of triplets were obtained from 8,500 subjects, each having the same 15 samples, the performance substantially improved to an EER of 1.12\%.  
    This is because, a model trained with data from 8,500 subjects covers a larger range of the feature space compared to a model trained with only 125 subjects. This expanded coverage contributes to the improved performance. Essentially, the better a model represents the entire spectrum of the features space, the better it performs.

    \item We observed a diminishing point of improvement, where further increasing the number of subjects had little to no impact on the performance, as evident in the comparison between results obtained from 8,500 and 68,000 subjects. 
    This observation shows that a model trained with 8,500 subjects probably already learned or mostly learned the entire feature space.
    Consequently, the subsequent addition of subjects had limited or negligible impact on its performance, explaining the diminishing returns observed.
    Hence, we determined that, for this specific dataset, 8,500 subjects represents an optimal number for effectively training the SNN with the Aalto dataset. 
\end{enumerate}

These findings underscore the significance of a broader training dataset in achieving significant performance improvements.
Note that our result (1.09\% EER) for 68,000 subjects in Table \ref{aalto_breadth_wise_table}, while showing a slight improvement, is comparable to the reported result (1.2\% EER) for the same number of subject in Acien et al. \cite{typenet}, further validating our experimental procedure.
\begin{table*}[tb]
\centering
\scriptsize
\caption{Aalto Dataset: The average EER for the breadth-wise experiments with 7.6 Million triplets, 15 samples per subject, and  sequence length 70.}
\label{aalto_breadth_wise_table}
\resizebox{\linewidth}{!}{
\begin{tabular}{c c c c c c c c c c c}
\toprule
& \multicolumn{10}{c}{\textbf{Group of Subjects}} \\
& \textbf{125} & \textbf{250} & \textbf{500} & \textbf{1K} & \textbf{2K} & \textbf{4K} & \textbf{8.5K} & \textbf{17K} & \textbf{34K} & \textbf{68K}\\
\toprule
\textbf{EER} & 7.94 & 4.99 & 2.91 & 1.82 & 1.37 & 1.21 & 1.12 & 1.12 & 1.11 & 1.09\\
\bottomrule
\end{tabular}
}
\end{table*}
\begin{figure}[htbp]
\begin{center}
\includegraphics[width=0.5\linewidth]{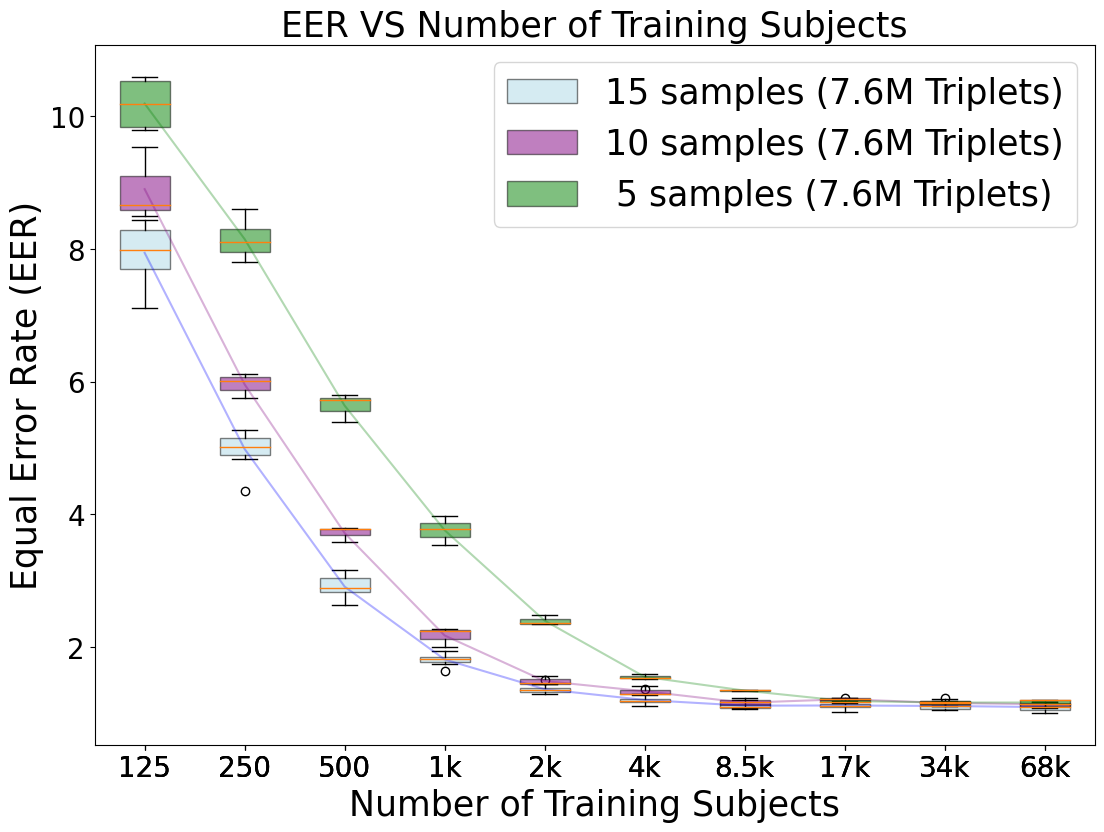}
\end{center}
   \caption{Aalto Dataset: Box plots for both the breadth-wise experiments (as seen horizontally with varying number of subjects), and the depth-wise experiments (as seen vertically with varying number of samples per subject) for 7.6 million triplets, where $M=70$. Each box plot displays the EERs from ten reruns.}
\label{aalto_depth_fig}
\end{figure}
\begin{table*}[t]
    \centering
    \begin{minipage}{.75\linewidth}
    \caption{Aalto Dataset: Average EERs for the depth-wise experiments with varying samples per subject, sequence length, and gallery sample size.}
    \label{aalto_depth_table1}
        \resizebox{\linewidth}{!}{
        \begin{tabular}{c | c | c | c c c c c c c c c c }
        \toprule
         & \textbf{Seq} & \textbf{Samples} & \multicolumn{10}{c}{\textbf{Average EER (\%) for Varying Number of Subjects with 7.6M Triplets}}\\
        \textbf{G} & \textbf{Len ($M$)} &
        \textbf{Per Subj} & \textbf{125} & \textbf{250} & \textbf{500} & \textbf{1K} & \textbf{2K} & \textbf{4K} & \textbf{8.5K} & \textbf{17K} & \textbf{34K} & \textbf{68K}\\
        \toprule
        \multirow{9}{*}{5} & \multirow{3}{*}{50} & 5 & 13.37 & 10.13 & 8.11 & 5.03 & 3.51 & 2.51 & 1.95 & 1.91 & 1.89 & 1.89 \\
        
        & & 10 & 12.68 & 8.29 & 5.05 & 4.04 & 2.25 & 1.97 & 1.88 & 1.84 & 1.78 & 1.77 \\
        
        & & 15 & 10.10 & 6.89 & 4.65 & 3.27 & 2.07 & 1.89 & 1.76 & 1.73 & 1.72 & 1.69 \\
        
        \cline{2-13}
        & \multirow{3}{*}{70} & 5 & 11.50 & 9.32 & 7.26 & 4.89 & 3.33 & 2.34 & 1.78 & 1.75 & 1.73 & 1.72 \\
        
        & & 10 & 10.13 & 7.05 & 4.17 & 3.02 & 2.05 & 1.89 & 1.66 & 1.64 & 1.61 & 1.61 \\
        
        &  & 15 & 9.05 & 5.52 & 3.75 & 2.93 & 1.91 & 1.66 & 1.65 & 1.63 & 1.61 & 1.55 \\
        
        \cline{2-13}
        & \multirow{3}{*}{150} & 5 & 10.93 & 8.81 & 5.21 & 2.97 & 2.83 & 2.07 & 1.69 & 1.67 & 1.65 & 1.63\\
        
        & & 10 & 9.16 & 6.72 & 3.72 & 2.90 & 2.00 & 1.81 & 1.65 & 1.63 & 1.61 & 1.59 \\
        
        &  & 15 & 8.75 & 5.50 & 3.65 & 2.85 & 1.90 & 1.65 & 1.63 & 1.62 & 1.60 & 1.53 \\
        
        \toprule
        \multirow{9}{*}{7} & \multirow{3}{*}{50} & 5 & 12.11 & 9.83 & 7.58 & 4.77 & 2.97 & 2.22 & 1.65 & 1.62 & 1.62 & 1.61 \\
        
        & & 10 & 11.12 & 7.89 & 4.63 & 3.21 & 2.38 & 1.84 & 1.63 & 1.58 & 1.57 & 1.57 \\
        
        & & 15 & 9.35 & 5.79 & 3.24 & 3.10 & 1.91 & 1.69 & 1.55 & 1.45 & 1.45 & 1.42 \\
        
        \cline{2-13}
        & \multirow{3}{*}{70} & 5 & 10.85 & 8.86 & 6.84 & 4.52 & 2.88 & 2.04 & 1.58 & 1.47 & 1.45 & 1.44 \\
        
        & & 10 & 9.71 & 6.46 & 3.98 & 2.59 & 1.80 & 1.63 & 1.47 & 1.45 & 1.44 & 1.41 \\
        
        &  & 15 & 8.40 & 4.99 & 3.32 & 2.33 & 1.74 & 1.51 & 1.46 & 1.44 & 1.44 & 1.37 \\
        
        \cline{2-13}
        & \multirow{3}{*}{150} & 5 & 8.42 & 7.01 & 4.71 & 2.55 & 2.36 & 1.47 & 1.45 & 1.41 & 1.35 & 1.38 \\
        
        & & 10 & 8.25 & 5.72 & 3.25 & 2.42 & 1.57 & 1.44 & 1.42 & 1.38 & 1.33 & 1.28 \\
        
        &  & 15 & 8.05 & 4.54 & 3.13 & 2.30 & 1.51 & 1.40 & 1.39 & 1.32 & 1.27 & 1.26 \\
        
        \toprule
        \multirow{9}{*}{10} & \multirow{3}{*}{50} & 5 & 10.68 & 8.48 & 6.25 & 3.87 & 2.73 & 1.61 & 1.42 & 1.36 & 1.30 & 1.27 \\
        
        & & 10 & 9.76 & 6.39 & 4.00 & 2.45 & 1.89 & 1.42 & 1.23 & 1.23 & 1.22 & 1.19 \\
        
        & & 15 & 8.39 & 5.34 & 3.13 & 1.93 & 1.49 & 1.34 & 1.19 & 1.18 & 1.18 & 1.18 \\
        
        \cline{2-13}
        & \multirow{3}{*}{70} & 5 & 10.19 & 8.15 & 5.64 & 3.76 & 2.40 & 1.55 & 1.35 & 1.21 & 1.16 & 1.11 \\
        
        & & 10 & 8.90 & 5.96 & 3.72 & 2.17 & 1.49 & 1.34 & 1.16 & 1.16 & 1.16 & 1.10 \\
        
        &  & 15 & 7.94 & 4.59 & 2.91 & 1.82 & 1.37 & 1.21 & 1.12 & 1.12 & 1.11 & 1.09 \\
        
        \cline{2-13}
        & \multirow{3}{*}{150} & 5 & 7.36 & 6.15 & 4.53 & 2.46 & 2.22 & 1.38 & 1.30 & 1.20 & 1.15 & 1.11 \\
        
        & & 10 & 7.11 & 5.12 & 3.44 & 1.98 & 1.42 & 1.30 & 1.15 & 1.14 & 1.14 & 1.10 \\
        
        &  & 15 & 7.01 & 4.04 & 2.65 & 1.76 & 1.30 & 1.20 & 1.10 & 1.11 & 1.11 & 1.10 \\
        \bottomrule
        \end{tabular}
        }
    \end{minipage}
    \begin{minipage}{.20\linewidth}
    \caption{Aalto Dataset: Average EERs for the depth-wise experiments with varying amount of training triplets. $G=10, M=70$ and 15 samples per subject}
    \label{aalto_depth_table2}
        \resizebox{\linewidth}{!}{
        \begin{tabular}{c | c | c}
        \toprule
        \textbf{Amount of} & \textbf{Selected} & \textbf{} \\
        \textbf{Triplets} & \textbf{Subjects} & \textbf{EER} \\
        \toprule
        \multirow{3}{*}{120K} & 125 & 7.97 \\
        & 4K & 2.35 \\
        & 68K & 2.17 \\
        \toprule
        \multirow{3}{*}{1M} & 125 & 7.95 \\
        & 4K & 1.38 \\
        & 68K & 1.21 \\
        \toprule
        \multirow{3}{*}{7.6M} & 125 & 7.94 \\
        & 4K & 1.21 \\
        & 68K & 1.09 \\
        \bottomrule
        \end{tabular}
        }
    \end{minipage}
\end{table*}

\subsection{Results of Depth-wise Experiments }
The impact of dataset depth is investigated using all three datasets. Note that while both Clarkson II and CMU are sufficiently deep, the depth of the Aalto is limited.

\subsubsection{Aalto Dataset} The results for the depth-wise experiments with the Aalto dataset are presented in Tables \ref{aalto_depth_table1} and \ref{aalto_depth_table2}, as well as Figure \ref{aalto_depth_fig}, highlighting the impact of dataset depth on the performance of SNN. The key observations are as follows:
\begin{enumerate}
    \item With an increase in the number of samples per subject from 5 to 10 and further to 15, we observed an improvement in performance (see Table \ref{aalto_depth_table1}).
    However, it is worth noting that while these performance improvements are more pronounced for models trained with triplets from a smaller pool of subjects (such as 125, 250 or 500 subjects), their significance diminishes and disappears as the number of subjects increases as shown in Table \ref{aalto_depth_table1}. This suggests that the influence of number of samples becomes less substantial after the number of subjects surpasses a certain threshold, which can be observed at 8,500 subjects. 
    This observation shows that increasing the number of samples is useful particularly when the  model is not yet well-trained and thus adding more samples contributes to both feature space and feature density.
    For instance, with 8,500 subjects, 
    the model is closer to being well-trained,
    so further increasing the number of samples per subject would have little to no effect on the performance.

    \item Table \ref{aalto_depth_table1} depicts a consistent performance increase as the sequence length and gallery sample size increase, showing that the model was effectively trained with regards to its depth.

    \item Recall that each experiment was repeated ten times, each time with a different subject selection, to ensure the reliability of our results. The box plots in Figure \ref{aalto_depth_fig} provide a visual representation of the EERs from these rerun experiments. Notably, we observed that the interquartile range within these box plots widens as the number of subjects gets smaller and  narrows as the number of subjects increases. This is because, with a limited number of subjects, the resulting learned feature space is less representative of the entire feature space (Recall the Aalto dataset was based on 1,525 sentences~\cite{dhakal2018observations}, which would comprise the scope of the feature space). Consequently, each rerun exhibits slight performance variations. However, when the dataset includes a larger number of subjects, the learned feature space becomes more representative, leading to consistently stable performance, depicting a sufficiently trained model.

    \item Furthermore, as shown in Table \ref{aalto_depth_table2}, generating 120K triplets versus 7.6M triplets from 125 subjects resulted in a negligible performance improvement (from 7.97 to 7.94 EER). However, repeating the same for 68K subjects, the performance improvement becomes notable (from 2.17 to 1.09 EER). This shows that the effect of varying the amount of training triplets is more pronounced on larger pool of subjects than on smaller ones. That is, as the subjects increase, the feature space to be learned also increases. As the number of subjects grows, so does the feature space to be learned. Training with larger triplets in such instances allows the model to expand its learned feature space, consequently leading to improved performance.
    This observation supports our hypothesis indicating that larger feature spaces, as encountered in larger subject pools, require a greater number of training triplets to comprehensively cover the entire feature space for effective training. As the quantity of training triplets increases, performance sees improvement until it reaches a plateau.
\end{enumerate}

\begin{table*}[tb]
\centering
\scriptsize
\caption{CMU Dataset: Average EERs for the depth-wise experiments with 7.6 million triplets.}
\label{cmu_depth_table}
\begin{tabular}{c | c | c c c c c}
\toprule
\textbf{Sequence} & \textbf{Samples} & \multicolumn{5}{c}{\textbf{Gallery (G)}} \\
\cline{3-7}
\textbf{Length ($M$)} & \textbf{Per Subject} & \textbf{5} & \textbf{10} & \textbf{20} & \textbf{40} & \textbf{60} \\
\toprule
\multirow{3}{*}{10} & 200 & 5.98 & 3.03 & 1.48 & 0.77 & 0.77 \\
 & 300 & 5.97 & 2.92 & 1.41 & 0.77 & 0.77 \\
 & 400 & 5.90 & 2.78 & 1.36 & 0.70 & 0.70 \\
\bottomrule
\end{tabular}
\end{table*}

\begin{table*}[tb]
\centering
  \scriptsize
    \caption{CMU Dataset: Average EERs for the depth-wise experiments with varying amount of training triplets. $G=10, M=10$ and 200 samples per subject}
    \label{cmu_depth_table2}
    \begin{tabular}{c c c c c}
    \toprule
    \textbf{} & \multicolumn{4}{c}{\textbf{Amount of Triplets}} \\
    \cline{2-5}
    \textbf{} & \textbf{120K} & \textbf{1M} & \textbf{7.6M} & \textbf{15.2M} \\
    \toprule
    \textbf{EER} & 3.04 & 3.02 & 3.03 & 3.00 \\
    \bottomrule
    \end{tabular}
\end{table*}

\subsubsection{CMU Dataset}
The results of the depth-wise experiments for the CMU dataset are presented in Table \ref{cmu_depth_table} and \ref{cmu_depth_table2}. The key findings include: 
\begin{enumerate}
    \item As shown in Table \ref{cmu_depth_table}, we observed a negligible performance improvement when the number of samples per subject was increased.
    Similarly, increasing the amount of training triplets used for training shows no significant performance improvement (3.04\% vs 3\% EER for 120K vs 15.2M triplets, respectively) as seen in Table \ref{cmu_depth_table2}. This aligns with our earlier hypothesis that the CMU dataset, being a fixed-text with a relatively fixed and smaller feature space, will not gain substantial benefit from increasing the samples per subject or amount of triplets. This is because, increasing the samples per subject or training triplets does not significantly expand the feature space but instead increases the already saturated density. Hence, this suggests that, in terms of dataset depth, a configuration with 200 samples per subject and 120K triplets is sufficient for effectively training the fixed-text CMU dataset.

    \item Furthermore, a significant performance improvement was observed (see Table \ref{cmu_depth_table}) as the gallery sample size increased, with the EER decreasing from 5.98\% at $G=5$ to 0.77\% at $G=40$. This performance improvement plateaued at $G=40$, with further increases in $G$ having no effect, which is consistent with our hypothesis of a sufficiently trained model.
    Notably, due to the CMU dataset's nature as a fixed-text, password-based dataset with a static and considerably smaller feature space, its performance significantly outperformed that of free-text datasets like the Aalto or Clarkson II, even after maintaining a small sequence length of 10 characters.
\end{enumerate}

Deng et al. \cite{deng2013keystroke} and Maheshwary et al. \cite{maheshwary2017deep} previously held the state-of-the-art records for the CMU dataset with EERs of 3.7\% and 3\%, respectively, using Deep Belief Nets and a feed-forward neural network model. These methods employed binary classifiers trained with the first 200 samples of each subject's data, and require a separate model for each subject. However, our model
outperformed their results with a new state-of-the-art performance, achieving an impressively low EER of 0.7\% with just 40 gallery samples. This marked a significant 76\% improvement in performance.

\begin{table*}[t]
\centering
    \begin{minipage}{.55\linewidth}
      \caption{Clarkson II Dataset: Average EERs for the depth-wise experiments with 7.6 million triplets.}
      \label{cuii_depth_table1}
      \centering
      \resizebox{\linewidth}{!}{
        \begin{tabular}{c | c | c c c c c c}
        \toprule
        \textbf{Sequence} & \textbf{Samples} & \multicolumn{5}{c}{\textbf{Gallery (G)}} \\
        \cline{3-8}
        \textbf{Length ($M$)} & \textbf{Per Subject} & \textbf{10} & \textbf{20} & \textbf{30} & \textbf{40} & \textbf{50} & \textbf{60} \\
        \toprule
        \multirow{3}{*}{70} & 50 & 17.39 & 17.21 & 17.09 & 16.72 & 17.88 & 17.92 \\
         & 65 & 15.55 & 16.71 & 14.24 & 15.17 & 15.98 & 16.49 \\
         & 140 & 13.69 & 10.91 & 10.75 & 12.03 & 12.25 & 12.60 \\
         \toprule
        \multirow{3}{*}{150} & 50 & 11.13 & 11.72 & 10.93 & 11.98 & 12.02 & 12.15 \\
         & 65 & 10.67 & 10.99 & 11.74 & 11.52 & 11.10 & 10.49 \\
         & 140 & 10.61 & 9.53 & 10.25 & 10.00 & 8.97 & 10.04 \\
         \toprule
        \multirow{3}{*}{200} & 50 & 9.88 & 10.77 & 10.88 & 10.90 & 11.01 & 11.14 \\
         & 65 & 9.21 & 10.44 & 10.15 & 10.52 & 10.67 & 10.07 \\
         & 140 & 8.75 & 9.27 & 9.42 & 7.75 & 9.77 & 10.02 \\
        \bottomrule
        \end{tabular}
        }
    \end{minipage}%
    \hfill
    \begin{minipage}{.40\linewidth}
      \centering
      \scriptsize
        \caption{Clarkson II Dataset: Average EERs for the depth-wise experiments with varying amount of training triplets and Gallery. $M=200$, and 140 samples per subject}
        \label{cuii_depth_table2}
        \resizebox{\linewidth}{!}{
        \begin{tabular}{c | c c c c c c}
        \toprule
        \textbf{Amount of} & \multicolumn{6}{c}{\textbf{Gallery (G)}} \\
        \cline{2-7}
        \textbf{Triplets} & \textbf{10} & \textbf{20} & \textbf{30} & \textbf{40} & \textbf{50} & \textbf{60}\\
        \toprule
        \text{120K} & 15.11 & 14.19 & 15.33 & 12.25 & 14.69 & 14.92 \\
        \text{1M} & 12.89 & 11.29 & 12.76 & 9.38 & 10.97 & 12.17 \\
        \text{7.6M} & 8.75 & 9.27 & 9.42 & 7.75 & 9.77 & 10.02 \\
        \text{15.2M} & 8.72 & 8.89 & 9.13 & 6.02 & 9.32 & 9.88 \\
        \text{30.4M} & 8.66 & 8.67 & 8.94 & 7.21 & 8.93 & 9.72 \\
        \bottomrule
        \end{tabular}
        }
    \end{minipage} 
\end{table*}

\subsubsection{Clarkson II Dataset}
The results for the depth-wise experiments using the Clarkson II dataset are presented in Table \ref{cuii_depth_table1} and \ref{cuii_depth_table2}. The key findings are as follows:
\begin{enumerate}
    \item Analyzing the results in Table \ref{cuii_depth_table1}, we observed performance improvements as the number of samples per subject and sequence length increased, which aligns with our hypothesis.

    \item However, unlike other datasets used in our experiments, we noticed a unique pattern, characterized by performance fluctuation as the gallery sample size ($G$) increased. While this observation may seems contradictory to the ideal expectation, it does reflects the nature of the Clarkson II dataset, which is a fully uncontrolled free-text dataset, thus with a much larger feature space, intra-subject variance, and limited density. That is, adding more data to the gallery sample, in this case,
    introduces new data points that were not covered at training as a result of insufficient training triplets, which can negatively impact performance. This observation aligns with our hypothesis of an under-trained model.

    \item Based on the results from Table \ref{cuii_depth_table2}, we noticed a general trend of improved performance when the amount of training triplets is increased. However, the performance fluctuation persist as G increases, indicating that all models remained under-trained even with 30.4 million triplets.

\end{enumerate}


\subsection{Discussion}
Comparing the results of our study, it becomes evident that both dataset breadth and depth significantly impact the performance of deep learning SNN models. Notably, the findings emphasize the substantial role of dataset breadth in enhancing performance, particularly in capturing subject-specific nuances. Additionally, the concepts of feature space and density provide a lens through which the impact of dataset depth and    related parameters, such as the number of samples per subject, sequence length, training triplets, and gallery sample size, can be better comprehended, which rely mostly on the nature of the dataset being analyzed.



\section{Conclusion}
This study investigated the impact of dataset breadth and depth on the performance of a deep learning Siamese network model in the context of behavioral biometrics, specifically focusing on keystroke dynamics. The study utilized three datasets, which are the Aalto, CMU and Clarkson II datasets. We conducted breadth-wise and depth-wise experiments to evaluate the influence of dataset characteristics on model performance, and also used the concept of feature space and density to better understand the experiments. 
The results revealed that both dataset breadth and depth play crucial roles in the model's performance. When feasible, increasing the number of subjects involved in the training dataset had a significant positive impact on the model's performance, demonstrating the importance of capturing a wide range of behavioral patterns and accounting for inter-subject variability. On the other hand, when there are not enough subjects in a dataset, the dataset depth, characterized by the number of samples per subject, sequence length, and amount of training triplets also influenced performance.

For a fixed-text dataset such as the CMU dataset with a relatively fixed feature space, performance gains are expected when introducing more subjects to the dataset. However, when the samples per subject are already sufficiently large, increasing the number of samples per subject does not yield significant performance improvements.  Also, altering the sequence length may pose challenges due to the dataset's fixed nature.
In the case of a controlled free-text dataset like the Aalto dataset with a more extensive feature space and large density, performance improvements are attainable by increasing the number of subjects, samples per subject, sequence length, and training triplets. On the other hand, for a completely uncontrolled, free-text dataset like the Clarkson II, characterized by an extensive feature space and limited density, to sufficiently train a model, the key requirements are probably a larger number of subjects and/or a notably higher amount of training triplets. In addition, other factors like increasing the number of samples per subject and sequence length can also be considered, as well as how breadth/depth affect generalizability. Future work evaluating the effect of breadth and depth with different architectures, such as that of TypeFormer \cite{stragapede2023typeformer}, on these same datasets could also better quantify these relationships for keystroke dynamics.

\section*{Acknowledgment}
This work was partially supported by NSF Awards CNS-1650503, TI-2122746, and OAC-2244049.

\bibliographystyle{apacite}
\bibliography{bibliography}
\end{document}